\documentclass[letterpaper]{article} 
\usepackage{aaai23}  
\usepackage{times}  
\usepackage{helvet}  
\usepackage{courier}  
\usepackage[hyphens]{url}  
\usepackage{graphicx} 
\usepackage{natbib}  
\usepackage{caption} 
\usepackage{algorithm}

\usepackage{amssymb}
\usepackage{amsmath}
\usepackage{lipsum}  
\usepackage{adjustbox}
\usepackage{booktabs}
\usepackage{multicol}
\usepackage{multirow}

\usepackage{enumitem}
\usepackage[noend]{algpseudocode}

\usepackage{graphicx}
\usepackage{subfig}
\usepackage{xcolor}

\newcommand{\gray}[1]{{\color{gray}{#1}} \addcontentsline{tdo}{todo}{#1} }
\algnewcommand\algorithmicforeach{\textbf{for each}}
\algdef{S}[FOR]{ForEach}[1]{\algorithmicforeach\ #1\ \algorithmicdo}

\usepackage{newfloat}
\usepackage{listings}
\DeclareCaptionStyle{ruled}{labelfont=normalfont,labelsep=colon,strut=off} 
\floatstyle{ruled}
\newfloat{listing}{tb}{lst}{}
\floatname{listing}{Listing}
\title{\textit{Learning From Drift}: \\Federated Learning on Non-IID Data via Drift Regularization}
\author{
   Yeachan Kim\textsuperscript{\rm 1},  Bonggun Shin\textsuperscript{\rm 2}
}
\affiliations{
    \textsuperscript{\rm 1}Deargen Inc., Seoul, South Korea\\
    \textsuperscript{\rm 2}Deargen USA Inc., Atlanta, GA\\
    \textsuperscript{\rm 1}yeachan@deargen.me, \textsuperscript{\rm 2}bonggun.shin@deargen.me

}

\usepackage{bibentry}

\begin{document}

\maketitle

\begin{abstract}

Federated learning algorithms perform reasonably well on independent and identically distributed (IID) data. They, on the other hand, suffer greatly from heterogeneous environments, i.e., Non-IID data. Despite the fact that many research projects have been done to address this issue, recent findings indicate that they are still sub-optimal when compared to training on IID data. In this work, we carefully analyze the existing methods in heterogeneous environments. Interestingly, we find that regularizing the classifier's outputs is quite effective in preventing performance degradation on Non-IID data. Motivated by this, we propose Learning from Drift (LfD), a novel method for effectively training the model in heterogeneous settings. Our scheme encapsulates two key components: drift estimation and drift regularization. Specifically, LfD first estimates how different the local model is from the global model (i.e., drift). The local model is then regularized such that it does not fall in the direction of the estimated drift. In the experiment, we evaluate each method through the lens of the five aspects of federated learning, i.e., Generalization, Heterogeneity, Scalability, Forgetting, and Efficiency. Comprehensive evaluation results clearly support the superiority of LfD in federated learning with Non-IID data.

\end{abstract}

\section{Introduction}
%

With the increasing privacy concerns, transmitting privacy-sensitive data (e.g., browsing history, electronic health records, or data containing intellectual property) to outside of local networks makes the training on different sources further difficult. To address the above challenge, federated learning enables multiple parties (i.e., regions, devices, and users) to cooperatively train a neural model without sending the local data between participants \cite{konevcny2016federated,mcmahan2017communication,yang2019federated,li2020federated}.

FedAvg \cite{mcmahan2017communication} is the most widely used method in practice. In this method, a centralized server distributes an initial model to participants, and they perform a local optimization on their local data. After the optimization, the trained models are aggregated on the server by averaging the trained parameters from different participants. However, as the local models\footnote{In this paper, we use the terms \textit{locally trained model} and \textit{local model} interchangeably. Similarly, we give no difference the terms between \textit{aggregated model} and \textit{global model}.} are trained to fit the local data distribution instead of the global data distribution, heterogeneity of data distribution (i.e., Non-IID data) degrades the performance of the federated learning \cite{li2018federated,li2019convergence,karimireddy2020scaffold,li2021model,li2021federated}. The promising way to overcome such degradation is to constrain the local optimization using the global model which is typically treated as more reliable and generalized than the local models \cite{li2021model}. However, recent works find that these approaches still suffer performance degradation and show no advantages over FedAvg in several heterogeneity settings \cite{luo2021no,li2021model}.

To further understand the performance degradation, we perform an in-depth analysis about the robustness against client drift when constraining outputs \cite{li2021model} and parameters \cite{li2018federated,luo2021no} of different layers on the settings of heterogeneous federated learning. Interestingly, we observe that regularizing the classifier's outputs, which is not the target in most works, is quite effective to prevent the drift while constraining others still suffer from the client drift. 


Motivated by our upfront analysis, we propose a novel method to prevent client drift in the heterogeneous environment, coined Learning from Drift (LfD). Unlike the previous works that the model is constrained to generate the same features and parameters as the global model, our scheme mainly focuses on the prediction differences on the same data between the local and the global model (we denote the difference as \textit{drift}), and it roughly contains how the local model differs from the global model over the categorical distribution. Based on that, the model is trained with the drift regularization, which enforces the training model not to fall in the drift direction.

We compare LfD with strong baselines through the lens of five important factors for federated learning on heterogeneous environments, which are generalization, heterogeneity, scalability, catastrophic forgetting, and efficiency. Comprehensive results show that LfD effectively prevents client drift while yielding strong performance for federated learning. For example, compared to strong baselines, LfD achieves the state-of-the-art performance and shows nearly the same performance as the training on the total dataset in several heterogeneous settings. In summary, our contributions include followings:
\begin{itemize}
    \item We perform an in-depth analysis about the client drift when constraining different parts of the local model. Interestingly, we find that constraining classifier's output is most effective to prevent the client drift while others still suffer from the drift (Section \ref{motivation}).
    \item Based on our analysis, we propose a robust federated learning algorithm against diverse heterogeneous settings by explicitly estimating the drift and regularizing local models over the prediction space. (Section \ref{method}).
    \item We compare the proposed method with the strong baselines via the five important aspects of federated learning and observe that LfD achieves the best performance over diverse heterogeneous settings and datasets (Section \ref{experiment}).
\end{itemize}

\section{Problem setup}

\textbf{Federated Learning.} We assume there is a central server that can transmit and receive messages from $K$ client devices in federated learning. Each client $k \in [K]$ has its local data $\mathcal{D}_k$ which consists of $N_k$ training instances in the form of input features $\boldsymbol{x}$ and its corresponding label $y$. The objective for federated learning is to learn a single model that performs well on the distributed dataset without sharing the data between clients. To this end, the server first sends a global model parameterized by $\boldsymbol{\omega}_{t}$ to each client where subscript $t$ indicates the current communication round, and the clients update the received model by optimizing the following local objective:
\begin{equation}\label{eq:local_opt}
    \min_{\boldsymbol{\omega}_{t}^{k}}\mathop{\mathbb{E}} {}_{(\boldsymbol{x}, y) \sim \mathcal{D}_{k}}[\mathcal{L}(\boldsymbol{x}, y; \boldsymbol{\omega}_{t}^{k})]
\end{equation}
where $\mathcal{L}$ is the algorithm-dependent loss function, and $\boldsymbol{\omega}_{t}^{k}$ indicates the locally optimized model initialized by the global model $\boldsymbol{\omega}_{t}$. After the local optimization, the server aggregates the local models $\boldsymbol{\omega}_{t}^{i} \text{ where } i \in \{1, 2,...K\}$ to update the global model by weighted averaging the models:
\begin{equation}\label{eq:aggregation}
    \boldsymbol{\omega}_{t+1} = \sum_{k = 1}^{K} p_{k} \boldsymbol{\omega}_{t}^{k}
\end{equation}
where the weight $p_{k}$ is typically determined by the number of the local data over the entire dataset, i.e., $p_{k} = |\mathcal{D}_{k}|/\sum_{j=1}^{K}|\mathcal{D}_{j}|$. The server and participants repeat the above procedures until the the global model converges.

\begin{figure}[t]
\centering 
\subfloat{%
  \includegraphics[width=\linewidth]{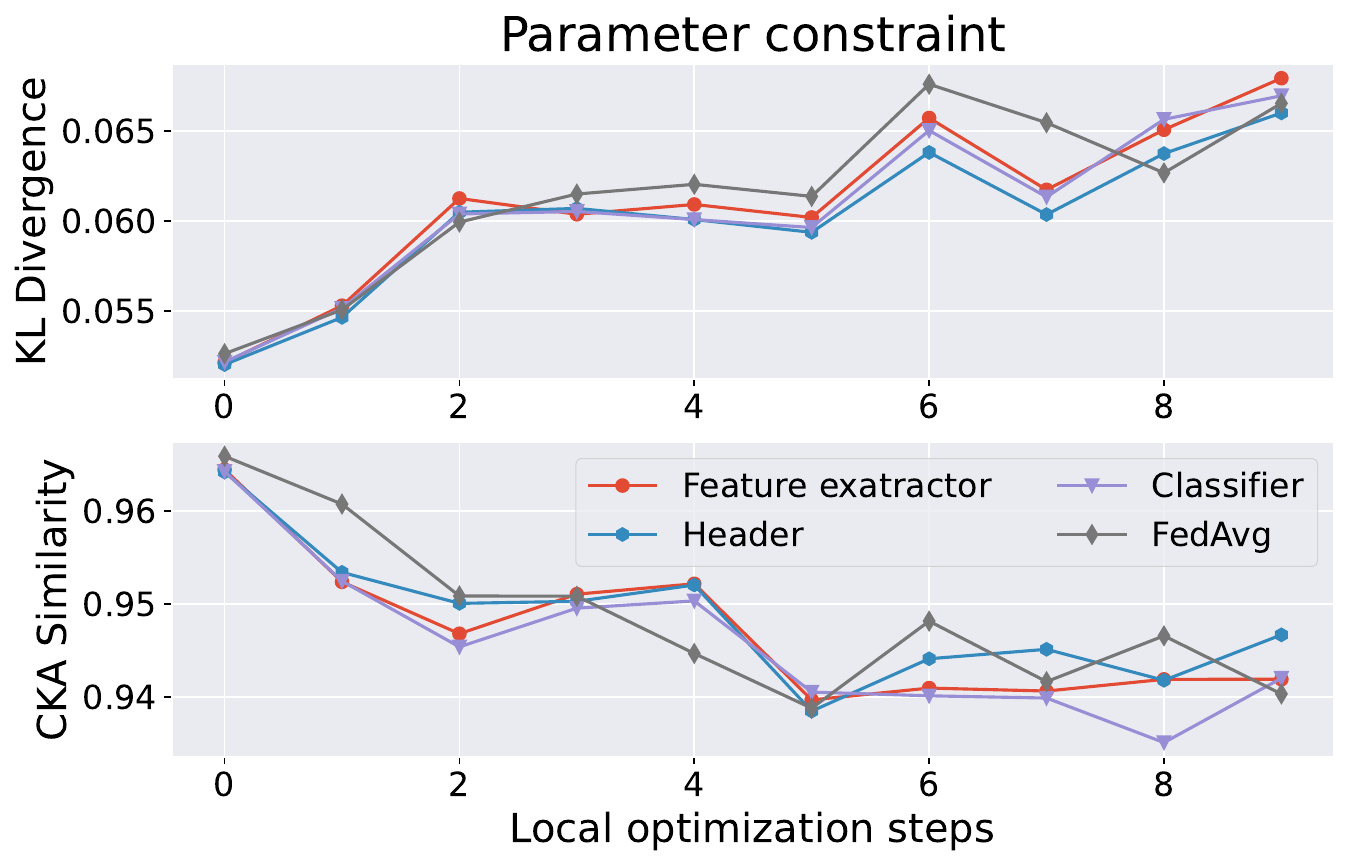}%
  \label{exp:in_forget} 
}

\subfloat{%
  \includegraphics[width=\linewidth]{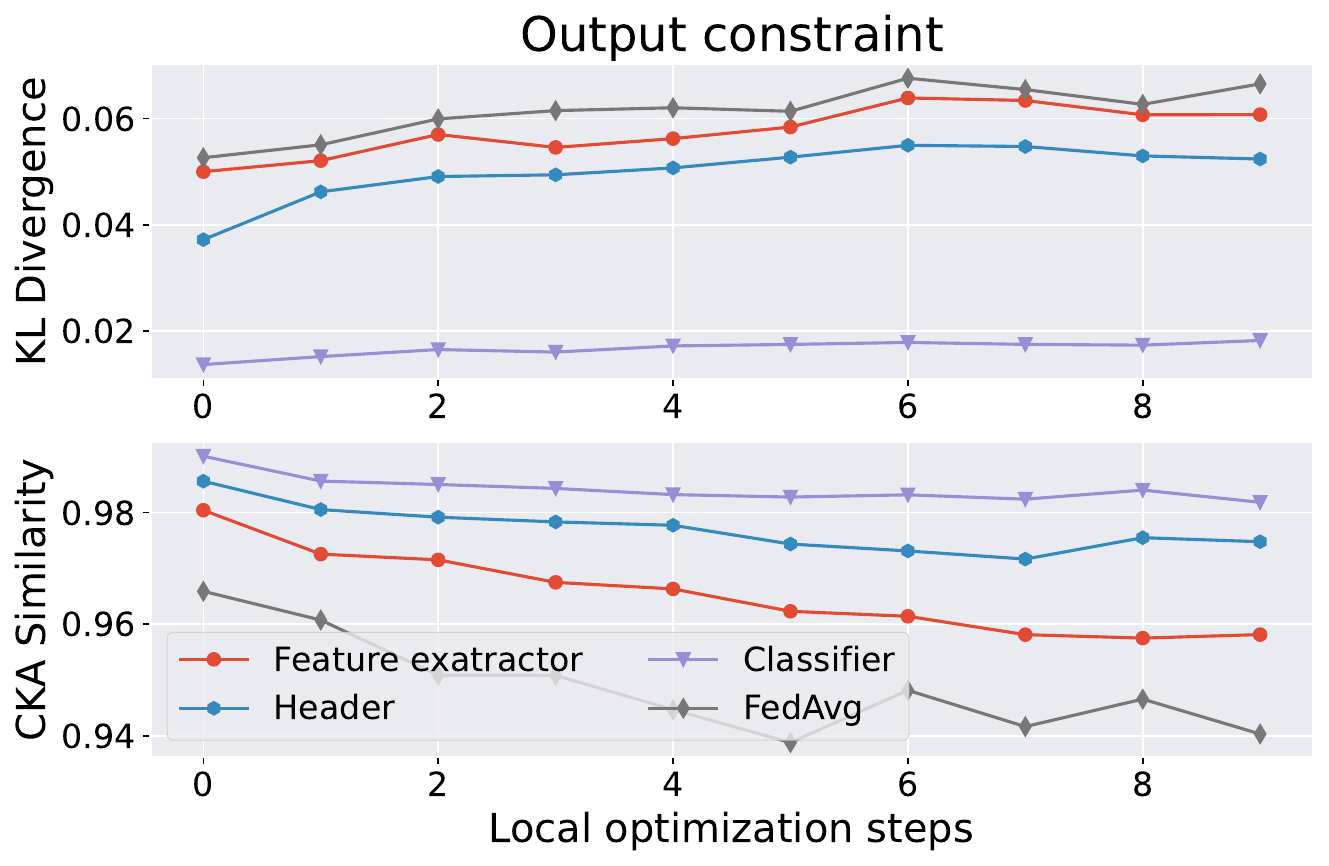}%
  \label{exp:out_forget}
}
\caption{CKA similarities and KL divergences during the local optimization with different global constraints.}
\label{fig:analysis}
\end{figure}



\section{Global constraint on Federated Learning}\label{motivation}

\begin{table}[t]
\caption{Top-1 Test accuracy (\%) of the methods with global constraints. The model without constraints is the same as FedAvg.}
\centering
\begin{adjustbox}{max width=\linewidth}
\begin{tabular}{@{}lcc@{}}
\toprule
Method     & \multicolumn{1}{c}{Feature constraint} & Parameter constraint \\ \midrule
No constraint     & \multicolumn{2}{c}{69.1 $\pm$ 0.6}                                      \\ \midrule \midrule
All                 & 67.3 $\pm$ 0.8 \small{\textcolor{purple}{($\downarrow$ 1.8)}}  & 68.6 $\pm$ 0.2 \small{\textcolor{purple}{($\downarrow$ 0.5)}}  \\
Feature extractor   & 68.9 $\pm$ 0.3 \small{\textcolor{purple}{($\downarrow$ 0.2)}}  & 69.5 $\pm$ 0.3 \small{\textcolor{teal}{($\uparrow$ 0.4)}}  \\
Header              & 69.6 $\pm$ 0.5 \small{\textcolor{teal}{($\uparrow$ 0.5)}}      & 69.3 $\pm$ 0.9 \small{\textcolor{teal}{($\uparrow$ 0.2)}}  \\
Classifier          & 70.8 $\pm$ 0.4 \small{\textcolor{teal}{($\uparrow$ 1.7)}}      & 69.7 $\pm$ 0.5 \small{\textcolor{teal}{($\uparrow$ 0.6)}}  \\ \bottomrule
\end{tabular}
\end{adjustbox}
\label{tab:motivation}
\end{table}

\begin{figure*}[ht!]
\centering
\begin{adjustbox}{max width=\linewidth}
    \includegraphics[width=0.85\textwidth,height=0.25\textheight]{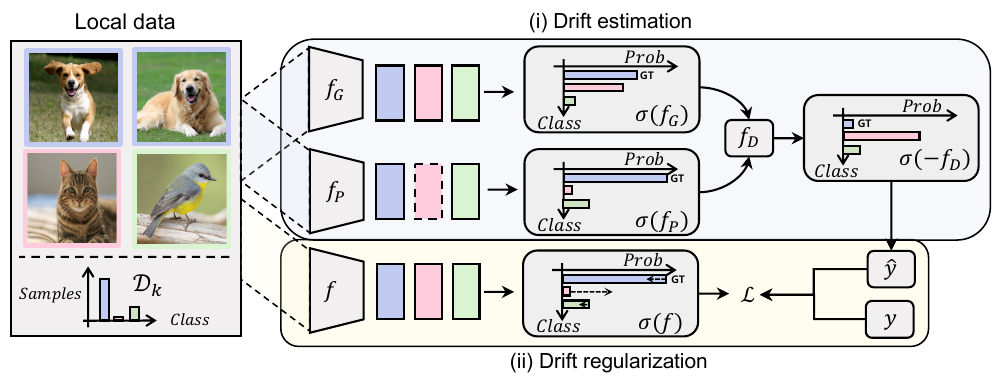}
    \label{model}
\end{adjustbox}
\caption{Overview of the proposed method (LfD). Here, the client has the imbalanced data distribution where the samples for \textit{dog} are the majority while the samples for \textit{cat} are the minority. In this case, LfD estimates that the client reveals over-confidence to the class \textit{dog} and under-confidence to the class \textit{cat}. Afterward, LfD regularizes the local model to predict the same input with slightly lower confidence for \textit{dog} and higher confidence for \textit{cat} so as to align with the global model.}
\label{fig:model}
\end{figure*}

One of the widely-used approaches to prevent client-drift is to constrain the model not to deviate from the global model. To understand how these constraints affect the training model in federated learning on Non-IID data, we perform an analysis by conducting experiments on heterogeneous environments. The setting includes ten clients with heterogeneously distributed CIFAR-10 data (see Section \ref{sect:simulation} for more information). Here, we constrain the model by minimizing the distance between the training local model and the global model, and the minimizing targets are feature vectors and parameters which are the representative targets in recent studies\footnote{For the feature constraint, the targets are the outputs from each component. For the parameter constraints, the targets are the trainable parameters of each layer.} \cite{li2018federated,li2021model,luo2021no}. The targets are further decomposed by their positions in the network architectures which are classifier, header\footnote{Header indicates the projection layers that are located between feature extractor and the classifier (i.e., the last layer). Headers are usually used at contrastive learning \cite{chen2020simple}.}, and feature extractor. We trace the output logits' similarity and prediction divergence to the global model during the local optimization to see the effects of the above constraints. For the similarity measure, we use centered kernel alignment (CKA) \cite{kornblith2019similarity} similarity and calculate the prediction divergence based on kullback-libeler divergence (KLD).

Figure~\ref{fig:analysis} reports the above measures during the local optimization. We first observe that the feature constraint encourages the training model to generate more similar features to the global model compared to the parameter constraint. When it comes to the divergence in the prediction, both constraints tend to show progressively deviating from the global model (i.e., increasing KL divergence) revealing that the model still suffers from the drift. However, it is noticeable that constraining the logits from the classifier makes the model less deviate from the global model, whereas regularizing the parameters of the classifier does not exhibit significant differences compared to other targets. 

As it is susceptible that constraining logits does not learn the local data while preserving the global knowledge, we estimate the federated learning performance of each constraint in Table \ref{tab:motivation}. From the evaluations, we first observe that the performance is more improved as regularization targets are deeper. In particular, we experimentally find that constraining the logits from the classifier is most effective to prevent the drift compared to other targets. Our finding is aligned with the recent work that demonstrates the deeper layers are vernerable to the client drift \cite{luo2021no}.

\section{Learning from Drift (LfD)}\label{method}

In this work, we propose a novel federated learning algorithm, coined Learning from Drift (LfD), for training a model in heterogeneous environments effectively. In an overview, LfD works in two major steps: drift estimation and drift regularization. In the first phase, LfD explicitly estimates the drift between the global model and the locally trained models on the logit space (Section~\ref{sect:define_drift}). Afterward, the model in each client is trained with the regularization, thereby leading the model not to fall in the drift direction during the local optimization (Section~\ref{sect:main_drift}). Figure~\ref{fig:model} shows the overview of our method.

\subsection{Drift Estimation by Prediction Discrepancy}\label{sect:define_drift}
Based on our upfront analysis, we quantify the degree of the client drift by estimating prediction discrepancy between locally trained models (i.e., $\boldsymbol{\omega}_{t-1}^{k}$) in the previous communication and their aggregated model (i.e., $\boldsymbol{\omega}_{t}$). The probability distribution over classes can be used to represent the prediction, and the drift for input $x$ is defined as follows:
\begin{equation}\label{eq:drift}
\begin{split}
    f^{y_i}_{D}(x) &= \log(\sigma(f^{y_i}_{P}(x))-\log(\sigma(f^{y_i}_{G}(x)) \text{ , } \forall y_i \in \mathcal{Y}
\end{split}    
\end{equation}
where $\sigma(\cdot)$ is the softmax function, and $f^{y_i}_{P}(\cdot), f^{y_i}_{G}(\cdot)$ are the logits of the local model trained in previous communication and its aggregated model (i.e., global model) for the class $y_i$, respectively. The estimated drift indicates how confident the local model is compared to the global model for each class. If the local model reveals more confidence\footnote{We define the confidence as the predicted probability to the specific class (e.g., ground-truth).} than the global model on the given input $x$, the large drift is estimated, and it can be interpreted as the local model was overfitted to the sample because the global model's confidence is more reliable and generalized than others \cite{li2021model}.



To prevent the drift, it is important to precisely estimate the drift by comparing the confidences between models. However, the confidences between the local models and the global model are not directly comparable due to the inconsistency between the magnitudes of the softmax weights and features. The softmax function for the prediction can be represented with the magnitudes and angles between input features and classifier weights.
\begin{equation}
\begin{split}
    \sigma(f^{y_i}(x;\omega)) &= \frac{e^{f^{y_i}(x)}}{\sum_{j=1}^{|\mathcal{Y}|}e^{f^{y_j}(x)}} = \frac{e^{W_{i}^{T}u}}{\sum_{j=1}^{|\mathcal{Y}|}e^{W_{j}^{T}u}}
    \\ &= \frac{e^{\Vert W_i\Vert \Vert u \Vert \cos\theta_i}}{\sum_{j=1}^{|\mathcal{Y}|}e^{\Vert W_j\Vert \Vert u \Vert \cos\theta_{ij}}},
\end{split}
\end{equation}
ewhere $\theta_{i}$ is the intersection angle between pre-activated features $u$ and classifier weights $W_i$ for the class $y_i$. As can be seen from the above equation, the magnitudes of features and the classifier weights can affect the confidence of the prediction by similarly serving as the different temperature. Different temperatures between classifiers (i.e., global and locals) result in different predictions in terms of the confidence, making the precise estimation of the drift difficult. Furthermore, the differences are more evident in the heterogeneous environments because this setting makes the classifier weights to be biased toward majority classes \cite{zhong2021improving}.

To estimate the drift precisely and comparably, we constrain the magnitudes by normalizing the classifier weights and features during the local optimization.
\begin{equation}
    \hat{u} = \frac{u}{\Vert u \Vert}, \hat{W_i} = \frac{W_i}{\Vert W_i \Vert}, \forall i \in \mathcal{Y},
\end{equation}
Based on the normalized features and classifier weights, the softmax function can be represented as:
\begin{equation}
    \sigma(f^{y_{i}}(x_{i};\omega)) = \frac{e^{\cos\theta_{i}}}{\sum_{k=1}^{|\mathcal{Y}|}e^{\cos\theta_{k}}},
\end{equation}
However, as the logit range is limited between zero to one (i.e., cosine range), the model can be slowly converged in the training. We thus give a margin to the ground-truth class to encourage the model to converge faster with the normalization constraint.
\begin{equation}
    \sigma(f^{y_{i}}(x_{i};\omega)) = \frac{e^{(\cos\theta_{i}-m)/\tau}}{e^{(\cos\theta_{i}-m)/\tau} + \sum_{k=1, k \neq i}^{|\mathcal{Y}|}e^{\cos\theta_{k}/\tau}},
\end{equation}
where $m \in [0,1]$ is the margin which is the hyper-parameter, $\tau$ indicates the temperature. The margin encourages the training model to predict the ground-truth label while having the margin $m$ to other classes on the logit\footnote{Note that we use the margin only in the training phase.}. This strategy allows to more correctly estimate the drift without slow convergence in the training.

\subsection{Drift Regularization}\label{sect:main_drift}

\begin{figure}[t]
\begin{adjustbox}{max width=\linewidth}
    \centering
    \includegraphics[width=1.0\textwidth]{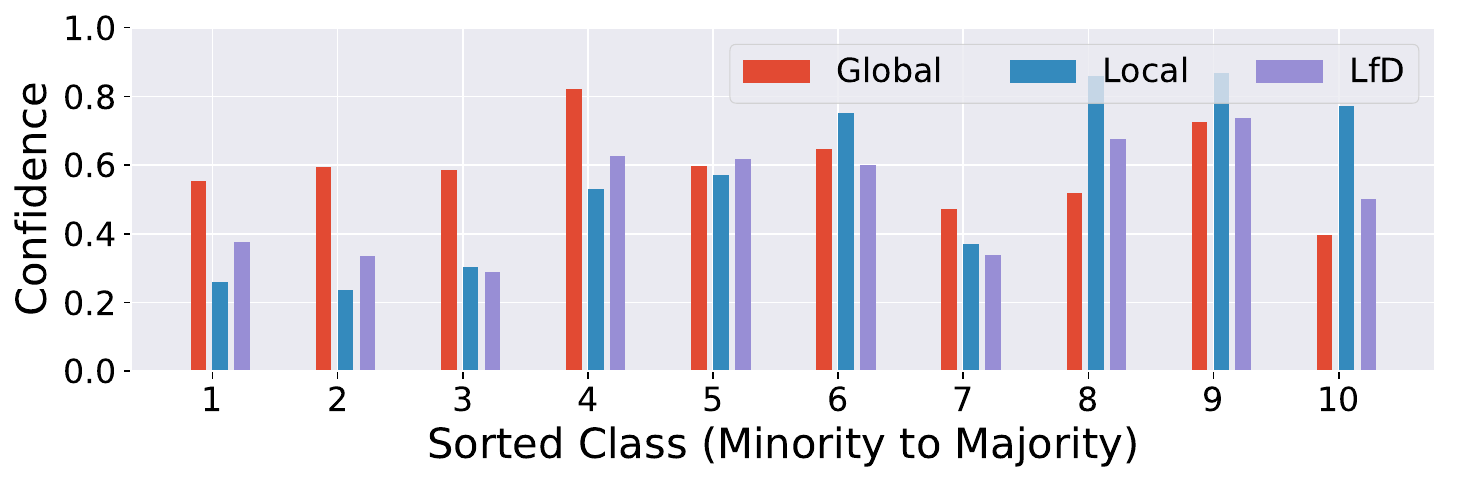}
    \label{model}
\end{adjustbox}
\caption{Confidence differences between global model, local model and local model with LfD. This shows that training with LfD encourages the local model to have the similar confidence to the global model.}
\label{fig:confidence_diff}
\end{figure}

After estimating the drift for each sample, the local optimization is performed such that the model does not fall in the drift direction. To this end, the model should be learned in the reverse direction to the estimated drift. We control the learning direction of the local optimization by regularizing the model with the auxiliary label containing the reverse direction of the estimated drift. We define the regularization term $\mathcal{R}$ to prevent the drift as follows:
\begin{multline}
    \mathcal{R}^{y_i}(x_i) = -\hat{y_i} \cdot \log(\sigma(f^{y_i}(x_i))) \text{ where } \hat{y_i} = \sigma(-f^{y_i}_{D}(x_i))
\end{multline}
We derive the reverse direction of the drift by taking the negative form of the estimated drift $f^{y_i}_{D}(x_i)$ and normalizing it by the softmax function. With this regularization, the local objective of each client can be formulated as:
\begin{equation}
    \mathcal{L}(\mathcal{D}, \omega) = -\sum_{(x, y) \in \mathcal{D}_k} \sum_{i = 1}^{|\mathcal{Y}|} (y_i\log\sigma(f^{y_i}(x)) + \mathcal{R}^{y_{i}}(x))
\end{equation}
where the first term is the cross-entropy with the ground-truth label, and the second term is the regularization with the auxiliary label. As the auxiliary label $\hat{y_i}$ includes the supervision for every class compared to the ground-truth, the objective can be decomposed according to the ground-truth and others.
\begin{equation}\label{eq:comb}
\mathcal{L}(x, \omega) = 
\begin{cases}
  -(1 + \hat{y_i}) \cdot \log\sigma(f^{y_{i}}(x)) &  \text{ if  } y_{i} = 1 \\
  - \hat{y_i} \cdot \log\sigma(f^{y_{i}}(x)) & \text{otherwise.}
\end{cases}
\end{equation}
For the ground-truth class, the regularization gives upward weights to the loss. The degree of the weights is determined by the confidence of the local model compared to the global model. The upward weights are set to low values if the local model reveals overconfidence. In contrast, higher upward weights are applied to samples for which the global model has higher confidence than the local model so as to avoid deviating from the global model. Figure \ref{fig:confidence_diff} shows the prediction confidence of the global model, the local model, and the local model regularized by LfD. It shows that the global and local models have different confidence, especially for the minority classes of the local data distribution. However, LfD effectively regularizes the local model to have similar confidence to the global model.

For the other classes, the regularization $\mathcal{R}$ works similarly to knowledge distillation \cite{hinton2015distilling} or label smoothing \cite{szegedy2016rethinking} by providing non-zero weights to other classes. It prevents the local model from forgetting inter-class similarities learned from the global model. This is particularly important when the client data distribution does not cover all classes in global data distribution. In the experiment, we show that the term prevents the model from forgetting the learned knowledge of the global model.

After the model is trained with LfD, the local models are sent to the central server for the aggregation (i.e., Eq.~\eqref{eq:aggregation}). We provide the overall process of LfD in Algorithm \ref{alg:lfd} in the supplementary material (Section \ref{sup:algo}).

\section{Evaluations}\label{experiment}

\subsection{Experimental Setup}
\subsubsection{Federated Simulation}\label{sect:simulation}
We mainly consider FedML benchmark \cite{he2020fedml} to evaluate each method on the federated learning scenario, \textit{i.e.,} CIFAR-10 \cite{krizhevsky2009learning}, CIFAR-100 \cite{krizhevsky2009learning}, and CINIC-10 \cite{darlow2018cinic}. We also validate our method on other domains to extend our knowledge to different domains, i.e., AGNews \cite{zhang2015character} for natural language processing and BindingDB \cite{liu2007bindingdb} for biology. To simulate federated learning, we randomly split the training samples of each dataset into $K$ batches (i.e., the number of clients, 10 by default), and assign one training batch to each client. As we are interested in Non-IID setting, we use Dirichlet distribution to generate the Non-IID data partition among participants. Specifically, we sample $p_{ik} \sim {Dir}_{K}(\beta)$ and allocate a $p_{ik}$ proportion of the instances of class $i$ to client $k$, where $Dir(\beta)$ is the Dirichlet distribution with a concentration parameter $\beta$ (0.5 by default), and smaller $\beta$ makes more heterogeneous data distribution.

\subsubsection{Baselines and Implementations}
We consider comparing the test accuracies of the representative federated learning algorithms: FedAvg \cite{mcmahan2017communication}, FedProx \cite{li2018federated}, FedAvgM \cite{hsu2020federated}, and state-of-the-art method MOON \cite{li2021model}. We carefully choose the best hyper-parameters used in FedProx (\textit{i.e.,} weighting factor $\mu$ for euclidean distance), FedAvgM  (\textit{i.e.,} $\beta$ momentum factor) and MOON (\textit{i.e.,} $\tau$ temperature and $\mu$ weighting factor for contrastive learning) through the validation set. For our method, we set the temperature as 0.1 (CIFAR-10, CINIC-10), 0.05 (CIFAR-100) for all softmax functions\footnote{We use the small temperature due to the normalization.} and set the margin as 0.15 by default. 

We use a simple 2-layers CNNs with 2-layers MLP projection networks for CIFAR-10. For CIFAR100 and CINIC-10, we adopt ResNet-18 \cite{he2016deep}. For a fair comparison, we use the same networks and augmentation for all methods. For the optimization, we use the SGD optimizer with a learning rate 0.01 for all approaches. The SGD weight decay is set to 0.00001 and the SGD momentum is set to 0.9. The batch size is set to 128. The number of local epochs is set to 300 for Union. The number of local epochs is set to 10 for all federated learning approaches unless explicitly specified. Each result is averaged result over three trials and these are implemented by PyTorch 1.7 on the NVIDIA A100 GPUs.

\begin{table}[t]
\caption{(Generalization) Top-1 Test Accuracy (\%) on CIFAR-10, CIFAR-100, BindingDB and AGNews. Best and second best results are highlighted in \textbf{boldface} and \underline{underlined}, respectively.}
\begin{adjustbox}{max width=\linewidth}
\begin{tabular}{@{}lcccc@{}}
\toprule
Method      & CIFAR-10 & CIFAR-100 & BindingDB & AGNews \\ \midrule
Union       & 74.5 \small{$\pm$ 0.9}    & 70.9 \small{$\pm$ 0.4}      & 89.5 \small{$\pm$ 1.2}       & 90.1 \small{$\pm$ 0.7}   \\ \midrule \midrule
FedAvg      & 69.1 \small{$\pm$ 0.6}    & 64.0 \small{$\pm$ 0.3}      & 85.7 \small{$\pm$ 0.8}       & 85.6 \small{$\pm$ 1.2}  \\
FedAvgM     & 69.9 \small{$\pm$ 1.1}    & 64.1 \small{$\pm$ 0.9}      & 86.5 \small{$\pm$ 1.5}       & 86.1 \small{$\pm$ 0.8}     \\
FedProx     & 68.6 \small{$\pm$ 0.9}    & 63.1 \small{$\pm$ 0.7}      & 86.1 \small{$\pm$ 0.5}       & \underline{86.9} \small{$\pm$ 0.5}     \\
MOON        & \underline{71.2} \small{$\pm$ 0.5} & \underline{64.3} \small{$\pm$ 0.8}      & \textbf{89.0} \small{$\pm$ 0.3}      & 85.4 \small{$\pm$ 0.4}\\
LfD (ours) & \textbf{74.1} \small{$\pm$ 0.4}      & \textbf{69.4} \small{$\pm$ 0.5} & \underline{88.1} \small{$\pm$ 0.4}       & \textbf{88.4} \small{$\pm$ 0.5}    \\ \bottomrule
\end{tabular}
\end{adjustbox}
\label{tab:generalization}
\end{table}

\subsection{Experimental Results}
We evaluate each method on following viewpoints to see the strength of the proposed methods in federated learning:
\begin{itemize}
    \item \textbf{Generalization}: The federated learning algorithm should be well generalized to diverse domains and datasets.
    
    \item \textbf{Heterogeneity}: The federated learning algorithm should be robust on different levels of heterogeneity, \textit{i.e.,} class skewness and training data distribution.
    
    \item \textbf{Scalability}: In reality, the number of clients is highly variable depending on the application. Moreover, there is no guarantee that all participants join the update steps due to communication loss or battery issues. Therefore, the federated learning method should work well with a large number of clients and the variation of the participating clients for the global update.
    
    \item \textbf{Forgetting}: One of the challenges for federated learning with Non-IID data is the class distribution's discrepancy, which interferes with the converge of the global model. To mitigate such a bad effect, the local model should maintain the knowledge learned from the global model during the local optimization.
    
    \item \textbf{Efficiency}: Since the communication between clients and server is the main source of energy consumption \cite{yadav2016review,latre2011survey}, it is important that federated learning methods should achieve strong accuracy with the small number of communications.
    
\end{itemize}

\subsubsection{Generalization}
We first evaluate whether each method can be generalized to different domains and datasets. We conduct the experiments on three different domains, i.e., image (CIFAR-10/100), drug discovery (BindingDB), and natural language (AGNews), and the evaluation results are tabulated in Table~\ref{tab:generalization}. It can be observed that the model learned by LfD yields strong accuracies on all domains (e.g., LfD achieves three state-of-the-art performances out of four domains), revealing that explicitly regularizing the local model based on the estimated drift is quite effective. Particularly, the models with global constraints, i.e., FedProx and MOON, often fail to achieve a great advantage over FedAvg. In contrast, LfD consistently shows improved accuracies by a large margin over FedAvg without considering domains and datasets. It further confirms the necessity of learning from the drift not just from the global constraints.

\subsubsection{Heterogeneity}
We confirm whether each method still works well with the different levels of heterogeneity and the shift of training data distribution. To this end, we adjust the concentration parameter $\beta$ to increase the level of heterogeneity (i.e, $\beta \in \{0.5, 0.1, 0.05\}$) and perform the experiment on CINIC-10 which is constructed from ImageNet \cite{russakovsky2015imagenet} and CIFAR10. As the samples from CINIC-10 are not drawn from an identical distribution, we can naturally evaluate how sensitive each method is to the distribution shift. We tabulate the evaluation results on Table~\ref{tab:heterogeneity}. We see that the state-of-the-art method, MOON, starts to largely lose accuracy as the level of heterogeneity increases. Specifically, when the local dataset is highly skewed (i.e., $\beta$ = 0.05), MOON shows roughly 10\% lower accuracy than FedAvg. Similar observation is observed in recent work \cite{luo2021no}. In contrast, LfD consistently yields higher accuracy than other baselines. For the case where the training distribution is shifted, i.e., CINIC-10, we can see LfD still works well, revealing that LfD is robust on both the different levels and types of heterogeneity.

\begin{table}[t]
\caption{(Heterogeneity) Top-1 Test Accuracy (\%) on CIFAR-10 with different level of heterogeneity ($\beta \in$ \{0.5, 0.1, 0.05\}), and CINIC-10. Best and second best results are highlighted in \textbf{boldface} and \underline{underlined}, respectively.}
\begin{adjustbox}{max width=\linewidth}
\begin{tabular}{@{}lcccc@{}}
\toprule
Method      & $\beta = 0.5$ & $\beta = 0.1$ & $\beta = 0.05$ & \multicolumn{1}{l}{CINIC-10} \\  \midrule
FedAvg      & 69.1 \small{$\pm$ 0.6}        & 62.4 \small{$\pm$ 1.1}    & 55.3 \small{$\pm$ 0.4}     & 74.2 \small{$\pm$ 1.9}\\
FedAvgM     & 69.9 \small{$\pm$ 1.1}        & \underline{63.4} \small{$\pm$ 1.3}    & \underline{55.9} \small{$\pm$ 0.9}     & 74.8 \small{$\pm$ 0.3}\\
FedProx     & 68.6 \small{$\pm$ 0.9}                & 63.1 \small{$\pm$ 0.4}    & 54.9 \small{$\pm$ 0.8}     & 73.9 \small{$\pm$ 1.5}\\
MOON        & \underline{71.2} \small{$\pm$ 0.5}   & 61.6 \small{$\pm$ 0.7}    & 50.9 \small{$\pm$ 2.1}     & \underline{77.1} \small{$\pm$ 1.7}\\
LfD (ours) & \textbf{74.1} \small{$\pm$ 0.4}   & \textbf{65.1} \small{$\pm$ 0.3}    & \textbf{58.2} \small{$\pm$ 0.6}     & \textbf{79.8} \small{$\pm$ 1.1}\\ \bottomrule
\end{tabular}
\end{adjustbox}
\label{tab:heterogeneity}
\end{table}

\subsubsection{Scalability}
We evaluate the scalability to verify that each method can be extended to the more realistic scenario, i.e., the more number of clients and the randomly active clients. For the evaluation, we increase the number of clients to 50 and 100, i.e., $ K \in \{50, 100\}$, while decreasing the active clients\footnote{We denote \textit{active client} as the clients who participate in the aggregation in Eq.~\eqref{eq:aggregation}.} to 25\% and 50\%. The overall results are shown in Table~\ref{tab:scalability}. It can be observed that increasing the number of clients and reducing the percentage of active clients degrades the performance of all methods because these adjustments can be one of the factors increasing the heterogeneity. However, LfD outperforms all baselines in all settings, demonstrating that Lfd performs well on the more realistic scenarios.

\begin{table}[t]
\caption{(Scalability) Top-1 Test Accuracy (\%) on CIFAR-10 with different number of participants ($|\mathcal{P}| \in$ \{10, 50, 100\}) and active clients in each communication($r \in \{0.25, 0.5\}$). Best and second best results are highlighted in \textbf{boldface} and \underline{underlined}, respectively.}
\centering
\begin{adjustbox}{max width=\linewidth}
\begin{tabular}{@{}lcccc@{}}
\toprule
\multicolumn{1}{c}{\multirow{2}{*}{Method}} & \multicolumn{2}{c}{$K$ = 50}                           & \multicolumn{2}{c}{$K$ = 100}                            \\ \cmidrule(l){2-5} 
\multicolumn{1}{c}{}                        & \multicolumn{1}{l}{$r$ = 0.25} & \multicolumn{1}{l}{$r$ = 0.5} & \multicolumn{1}{l}{$r$ = 0.25} & \multicolumn{1}{l}{$r$ = 0.5} \\ \midrule
FedAvg      & \underline{61.6} $\pm$ 1.5    & 62.6 $\pm$ 0.8            & 55.8 $\pm$ 2.3   & 58.3 $\pm$ 1.6 \\
FedAvgM     & 60.6 $\pm$ 0.8                & 61.5 $\pm$ 0.2            & \underline{59.5} $\pm$ 0.9   & \underline{61.2} $\pm$ 0.5 \\
FedProx     & 61.5 $\pm$ 1.3                & 62.1 $\pm$ 0.6            & 56.8 $\pm$ 1.7   & 59.8 $\pm$ 0.4 \\
MOON        & 61.4 $\pm$ 0.9                & \underline{63.2} $\pm$ 1.2    & 56.3 $\pm$ 2.7   & 59.5 $\pm$ 0.9 \\
LfD (ours) & \textbf{64.8} $\pm$ 0.6       &\textbf{ 66.2} $\pm$ 0.7  & \textbf{61.3} $\pm$ 1.7   & \textbf{62.7} $\pm$ 1.1                      \\ \bottomrule
\end{tabular}
\end{adjustbox}
\label{tab:scalability}
\end{table}

\subsubsection{Forgetting}
\begin{figure}[t]
\centering 
\subfloat{%
  \includegraphics[width=0.9\linewidth]{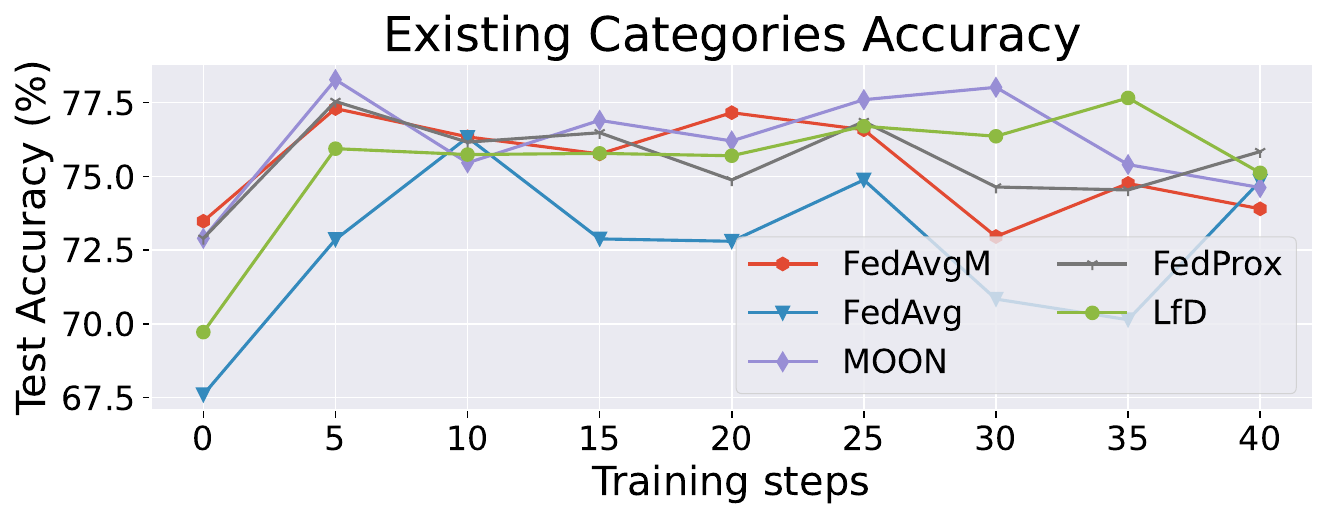}%
  \label{exp:in_forget} 
}

\subfloat{%
  \includegraphics[width=0.9\linewidth]{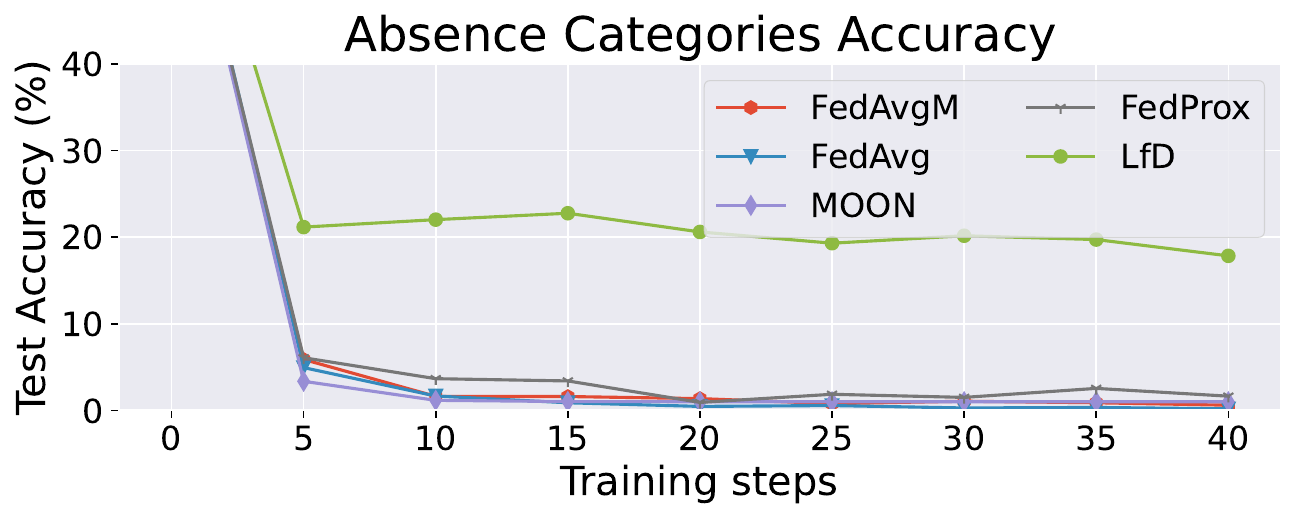}%
  \label{exp:out_forget}
}
\caption{Test accuracy (\%) for existing (Top) and absence (Bottom) classes during the local optimization. This shows that only LfD maintains the knowledge for the absence classes.}
\label{fig:forgetting}
\end{figure}

The global model usually has the better knowledge for the global distribution \cite{li2021model}. However, the learned knowledge can be forgotten after the local optimization as the training makes the model fit the local distribution. Such forgetting phenomena is called \textit{catastrophic forgetting}, and it is widely known that the forgetting leads to the slow convergence and worse performance on federated learning \cite{li2018federated,zhao2018federated}. To estimate how much each method forgets or maintains the learned knowledge, we calculate the accuracy for existing and absence classes before and after the local optimization and quantify the forgetting by defining learning performance (LP) as:
\begin{equation}
    LP(\mathcal{D}^{test};\omega_{t}^{k},\omega_{t}) = \frac{1}{|\mathcal{Y}|}\sum_{y_i \in \mathcal{Y}}\frac{Acc(\mathcal{D}^{test}(y_i), \omega_{t}^{k})}{Acc(\mathcal{D}^{test}(y_i),\omega_{t})}
\end{equation}
where $\mathcal{D}^{test}(y_i)$ is the test dataset that contains samples belonging the class $y_i$, and $Acc()$ is the accuracy on the given dataset and classifier. $LP(\cdot)$ is the relative metric estimating how the categorical accuracy is changed after the local optimization. For the case where $LP(\cdot)$ is larger than the value of one, we can interpret this as the improved accuracy after the optimization. In contrast, the lower value of one indicates the forgetting after the optimization. Based on this metric, we select the specific client who has the only half of categories and estimate $LP$ for existing and absence categories. Table~\ref{tab:forgetting} and Figure~\ref{fig:forgetting} shows the LP performance and its dynamics during the local optimization, respectively\footnote{Here, we use the trained models from each method that have nearly similar test accuracy to the global model.}. From the table, the accuracy for existing categories is improved after the optimization in all baselines. However, the baselines completely forget the discriminative ability for the absence classes. Interestingly, Lfd maintains the learned knowledge to some extent without learning any samples for the corresponding classes. The results can be explained by the distillation effect of the drift regularization on Eq.~\eqref{eq:drift}.

\begin{table}[ht]
\caption{(Forgetting) Learning performance on CIFAR10 for existing and absence classes.}
\centering
\begin{adjustbox}{max width=\linewidth}
\begin{tabular}{@{}lcc@{}}
\toprule
Method      & Existing categories & \multicolumn{1}{l}{Absence categories} \\ \midrule
FedAvg      & 1.06                 & 0.00  \\
FedAvgM     & 1.04                 & 0.00  \\
FedProx     & 1.05                & 0.03  \\
MOON        & 1.05                & 0.02  \\
LfD (ours) & 1.07                & 0.27  \\ \bottomrule
\end{tabular}
\end{adjustbox}
\label{tab:forgetting}
\end{table}

\begin{figure}[t]
\centering 
\subfloat{%
  \includegraphics[width=0.87\linewidth]{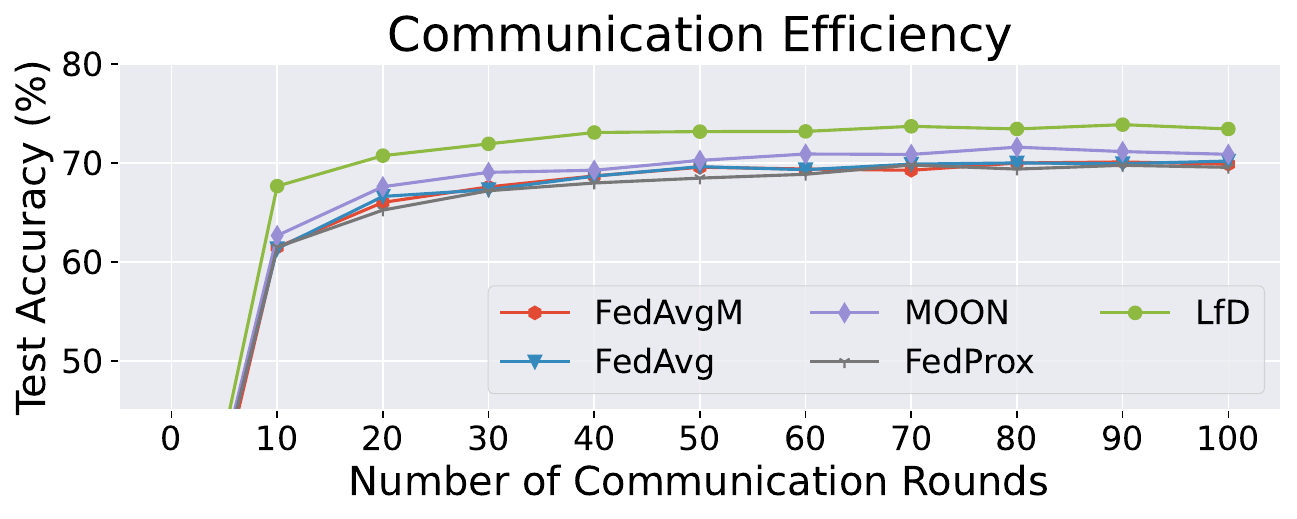}%
  \label{exp:comm_eff} 
}

\subfloat{%
  \includegraphics[width=0.87\linewidth]{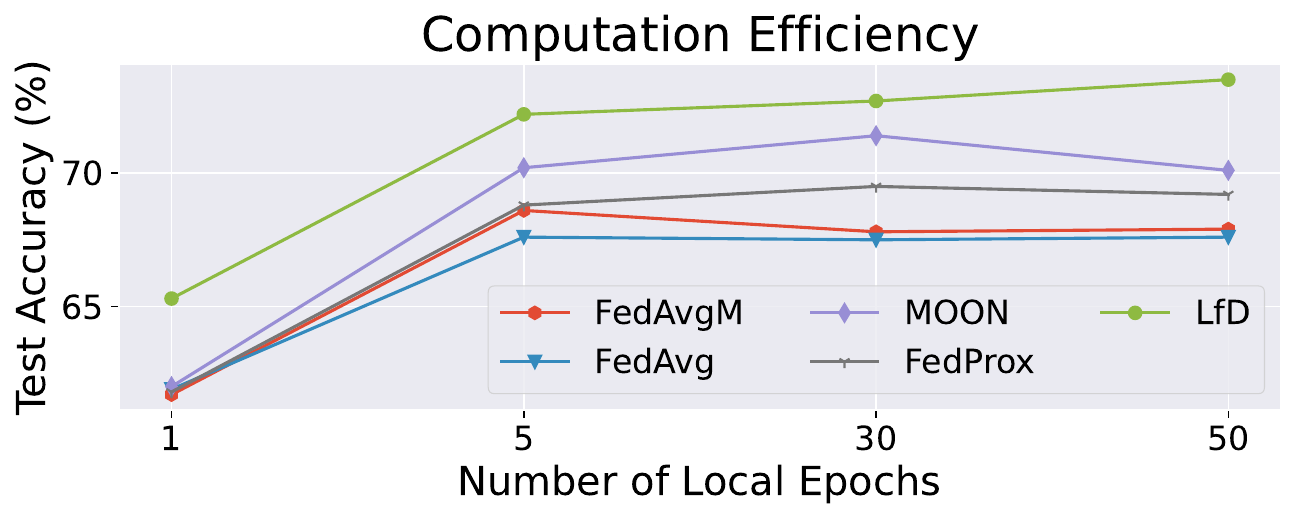}%
  \label{exp:comp_eff}
}
\caption{Test accuracy over the different number of communication rounds (Top) and local optimization steps (Bottom).}
\label{fig:eff}
\end{figure}

\subsubsection{Efficiency}
We analyze how the number of communication rounds and local optimizations affect the performance of the global model. The experimental settings are the same for the CIFAR-10 dataset. Figure \ref{fig:eff} shows the results. For communication efficiency, LfD reaches the best performance of other baselines in earlier rounds (e.g., LfD achieves 71.5\% on 22 rounds), indicating that the proposed method is less affected by the heterogeneous distributions. The analysis of computation efficiency could support the above results. LfD tends to obtain better performance on more local optimizations while other baselines start to lose the accuracy after between 5 and 30 epochs due to the client drift, which is commonly observed in other studies \cite{li2018federated,li2021model}. These results validate that LfD can achieve strong accuracy with the small number of communication and computation costs.

\section{Related work}
In this section, we mainly review the federated learning algorithms dealing with Non-IID data because it is one of the most challenging settings that degrade the performance and slow down the convergence \cite{li2018federated,li2021federated}.

\subsection{Global Regularization}
While the FedAvg method works well with IID data, the performance significantly degrades with increasing heterogeneity on data distribution \cite{li2018federated}. To prevent the worse effect, one of the mainstream is to give constraint to local optimization by maximizing agreement with the global model. SCAFFOLD \cite{karimireddy2020scaffold} corrects local optimization by introducing \textit{control variate} from the global model. FedProx \cite{li2018federated} regularize the networks by minimizing the euclidean distance to the global model. However, recent work \cite{li2021model} shows that those works have little or even no advantage over FedAvg and propose MOON which adopts the contrastive learning to increase agreement with the global model based on projected head vectors. FedDyn \cite{acar2021federated} changes the local optimization to ensure that the local optimum on each client is asymptotically consistent with the stationary points of the global objective.

\subsection{Model Aggregation}
Instead of naively averaging the model weights, some of the works revise the aggregation step in the orchestration server. FedAvgM \cite{hsu2020federated} updates the aggregated model with server momentum. FedMA \cite{wang2020federated} mitigate the model heterogeneity by matching similar neurons (e.g., convolutional filters and hidden states in LSTMs) between clients and performing the average to build the global model. Similarly, Fed$^2$ \cite{yu2021fed2} align similar features by model structure adaptation and feature-paired averaging on similar functioning neurons.

\subsection{Data Sharing and Generation}
Sharing the local data between clients is prohibited in the federated learning scenario. One of the practices is to share public data \cite{zhao2018federated,li2019fedmd} and unlabeled or synthesized data points \cite{lin2020ensemble}. Instead of sharing or generating input points, \cite{luo2021no} estimate the feature statistics and augment the features followed the global distribution to calibrate the classifier. These works are promising but have potential risks for privacy preservation.

\section{Conclusion}
In this work, we have analyzed the existing federated learning algorithms that employ the global model to prevent client drift. We have observed that regularizing the features on logit space is most effective, whereas constraining other features and parameters does not provide much improvement over FedAvg. Based on the upfront analysis, we propose a federated learning algorithm, coined Learning from Drift (LfD). LfD explicitly estimates the client drift over logit space and regularizes the local model to learn the inverse direction of the estimated drift. In the experiments, we have evaluated our method and strong baselines with regard to five perspectives (i.e., Generalization, Heterogeneity, Scalability, Forgetting, and Efficiency). Comprehensive evaluation results clearly show that LfD effectively prevents client drift and achieves state-of-the-art results on experiments with diverse heterogeneous settings.

\bibliography{aaai23}
\appendix

\clearpage
\begin{center}
\LARGE
\textbf{Supplementary Materials for LfD}    
\end{center}

\section{Algorithm for LfD}\label{sup:algo}

\begin{algorithm}[h]
\caption{Federated Learning with LfD}
\label{alg:lfd}
\begin{algorithmic}[1]
\Require Number of communication rounds $T$, Number of local optimization epochs $E$, Local learning rate $\eta$, Number of clients $K$ and their local data $\mathcal{D}_k$ $\forall k \in \{1,2,...,K\}$
\State \gray{// Server side:}

\For {t $\in \{1, 2,...,T\}$}
    \For {k $\in \{1, 2,...,K\}$} (\textbf{parallel process})
        \State send the global model $\omega_t$ to client $k$
        \State $\omega^{k}_t$ $\gets$ \textbf{LocalOptimization}($k$, $\omega_t$)
    \EndFor
    \State aggregate the local model to build the global model $\omega_{t+1}$
    \State ${\omega}_{t+1} \gets \sum_{k = 1}^{K} \frac{|\mathcal{D}_k|}{|\mathcal{D}|}\cdot \omega_{t}^{k}$
\EndFor
\State \gray{// Client side:}
\State \textbf{LocalOptimization}(k, $\omega_t$):
\State $\omega^{k}_t \gets \omega_{t}$
\For {e $\in \{1, 2,...,E\}$}
    \ForEach {batch $B = \{b_{x}, b_{y}\} \in \mathcal{D}_{k}$}
        \State \gray{// (1) Drift Estimation}
        \State $f_D(b_x) \gets \log(\sigma(f_{P}(b_x))-\log(\sigma(f_{G}(b_x))$
        \State $\hat{b}_y \gets \sigma(-f_D(b_x))$ 
        \State \gray{// (2) Drift Regularization}
        \State $\mathcal{R}(B) \gets CrossEntropy(f(b_x), \hat{b}_y)$
        \State $\mathcal{L}(B) \gets CrossEntropy(f(b_x), {b}_y)$
        \State $\omega_{t}^{k} \gets \omega_{t}^{k} - \eta \nabla_{\omega_{t}^{k}} (\mathcal{L}(B) + \mathcal{R}(B))$
    \EndFor
\EndFor
\textbf{return} the locally trained model $\omega^{k}_{t}$
\end{algorithmic}
\end{algorithm}

\end{document}